\documentclass{article}
\usepackage{spconf,amsmath,graphicx}
\usepackage{multirow} 
\usepackage{booktabs} 
\usepackage{array}
\usepackage[dvipsnames]{xcolor}
\newcolumntype{C}[1]{>{\centering\arraybackslash}p{#1}} % Centered column type

\title{Zero-shot Audio Topic Reranking using Large Language Models}
\name{Mengjie Qian, Rao Ma, Adian Liusie, Erfan Loweimi, Kate M. Knill, Mark J.F. Gales}
\address{ALTA Institute, Machine Intelligence Lab, Department of Engineering, Cambridge University, UK}
\begin{document}
%\ninept
%
\maketitle

\footnotetext[1]{This paper reports on research supported by EPSRC Project EP/V006223/1 (Multimodal Video Search by Examples) and Cambridge University Press \& Assessment, a department of The Chancellor, Masters, and Scholars of the University of Cambridge.}

\begin{abstract}
Multimodal Video Search by Examples (MVSE) investigates using video clips as the query term for information retrieval, rather than the more traditional text query. This enables far richer search modalities such as images, speaker, content, topic, and emotion. A key element for this process is highly rapid and flexible search to support large archives, which in MVSE is facilitated by representing video attributes with embeddings. This work aims to compensate for any performance loss from this rapid archive search by examining reranking approaches. In particular, zero-shot reranking methods using large language models (LLMs) are investigated as these are applicable to any video archive audio content. Performance is evaluated for topic-based retrieval on a publicly available video archive, the BBC Rewind corpus. Results demonstrate that reranking significantly improves retrieval ranking without requiring any task-specific in-domain training data. Furthermore, three sources of information (ASR transcriptions, automatic summaries and synopses) as input for LLM reranking were compared. To gain a deeper understanding and further insights into the performance differences and limitations of these text sources, we employ a fact-checking approach to analyse the information consistency among them.
\end{abstract}
\section{Introduction}
\label{sec:intro}
For multimedia information retrieval, the search is usually based on text keywords, queries, or a precise summary of the desired video content. An interesting alternative is to search for videos by examples, where users present a single video and the retrieval system finds relevant content present in the corpus, Multimodal Video Search by Examples (MVSE), see for example~\cite{wang2024mvrmlm,gabeur2020multi,rouditchenko2021avlnet,suris2018cross}. This approach offers highly flexible search that can support multiple modes of query, for example image, speaker~\cite{loweimi2024usefulness}, topic, content, or emotion information. %prefer using plural nouns
 
This study focuses on one specific aspect of MVSE, namely audio topic retrieval where the objective is to help users find audio with topics that belong to the same topic category (such as BusinessAndIndustry, Education, Health) as the audio of the selected clip. When meta-data is available, for example the title, synopsis, or user-uploaded captions, these can be used to support topic retrieval. However, for many videos this information is not available. Hence for this work, we focus on the general scenario where the retrieval system only relies on features that can be automatically derived from the video. Specifically, we generate transcriptions with state-of-the-art automatic speech recognition (ASR) models, Whisper~\cite{radford2022robust}, and focus on maximising the topic retrieval performance in this setup.

In this paper, we split the search process into two stages: 
\begin{itemize}
    \item Stage 1: Initial Embedding-based Retrieval
    \item Stage 2: LLM-based Reranking
\end{itemize}
In stage 1, each audio file is represented by vector embeddings to support very rapid retrieval for large archives~\cite{yang2023unsupervised}. 
Previous studies~\cite{reimers-2019-sentence-bert,nogueira-etal-2020-document,pradeep2021expando,liusie2022university} have shown the effectiveness of reranking, however, these works typically fine-tune the reranker on task-specific query-archive pairs,  relying on labelled data that is not always available or sufficiently general to support all archives and queries. With the impressive emergent ability of current large language models (LLMs) \cite{zheng2024judging,ouyang2022training, zhao2023survey}, zero-shot evaluation methods have demonstrated comparable performance that can often rival the performance of supervised methods \cite{ma2023zero, qin2023large, liusie2023zero}. In this work, there is no in-domain text data for training a reranking model. Therefore, we examine the impact of zero-shot reranking approaches for the audio topic retrieval task in Stage 2.

Experiments in this paper are based on audio extracted from public BBC video recordings. For each video in the dataset, a human-generated synopsis that concisely summaries the video is also provided. Our results show that a retrieval system built from ASR transcriptions yields inferior performance to one built from the provided human-written synopses, with an nDCG@3 of 0.47 compared to 0.61. %We note that this is expected, since synopses may include richer context information from the human annotators while also being more succinct and less noisy than the ASR transcripts. Although the retrieval system using synopses outperforms the one using ASR transcriptions, we successfully narrow the performance gap by adopting a second reranking stage. Results indicate that the zero-shot pairwise reranking method shows comparable performance to the listwise counterpart while being more cost-effective. With zero-shot pairwise reranking, the system performance is largely improved from 0.47 to 0.54 in terms of the nDCG@3 metric.
We hypothesise two main reasons for this performance difference: (1) synopses can include richer context information from human annotators given their in-domain expert knowledge, and (2) synopses are more succinct and less noisy than the ASR transcriptions. To test the first hypothesis, we adopt a few-shot fact-checking method to analyse the information consistency between synopses and ASR transcriptions. To address the second hypothesis, we propose using automatic summaries (AutoSum) of the ASR transcriptions, which are more concise and reduce redundancy. This approach successfully narrows the performance gap with the retrieval system using synopses, improving the nDCG@3 from 0.47 to 0.52. 
A second reranking stage, using either listwise  \cite{ma2023zero, sun2023chatgpt} or pairwise reranking \cite{qin2023large, liusie2023zero}, further reduces the performance gap. Results indicate that the zero-shot pairwise reranking method shows comparable performance to the listwise counterpart while being more computationally cost-effective. Pairwise reranking with AutoSum improved nDCG@3 to 0.54, representing a 14.9\% relative improvement compared to the baseline retrieval with ASR transcriptions.

\section{MVSE Audio Topic Retrieval}
\subsection{Baseline Retrieval System}
\label{sec:mvse}
\begin{figure}[!t]
    \centering
    \includegraphics[trim=0 10mm 0 0mm, width=1.0\linewidth]{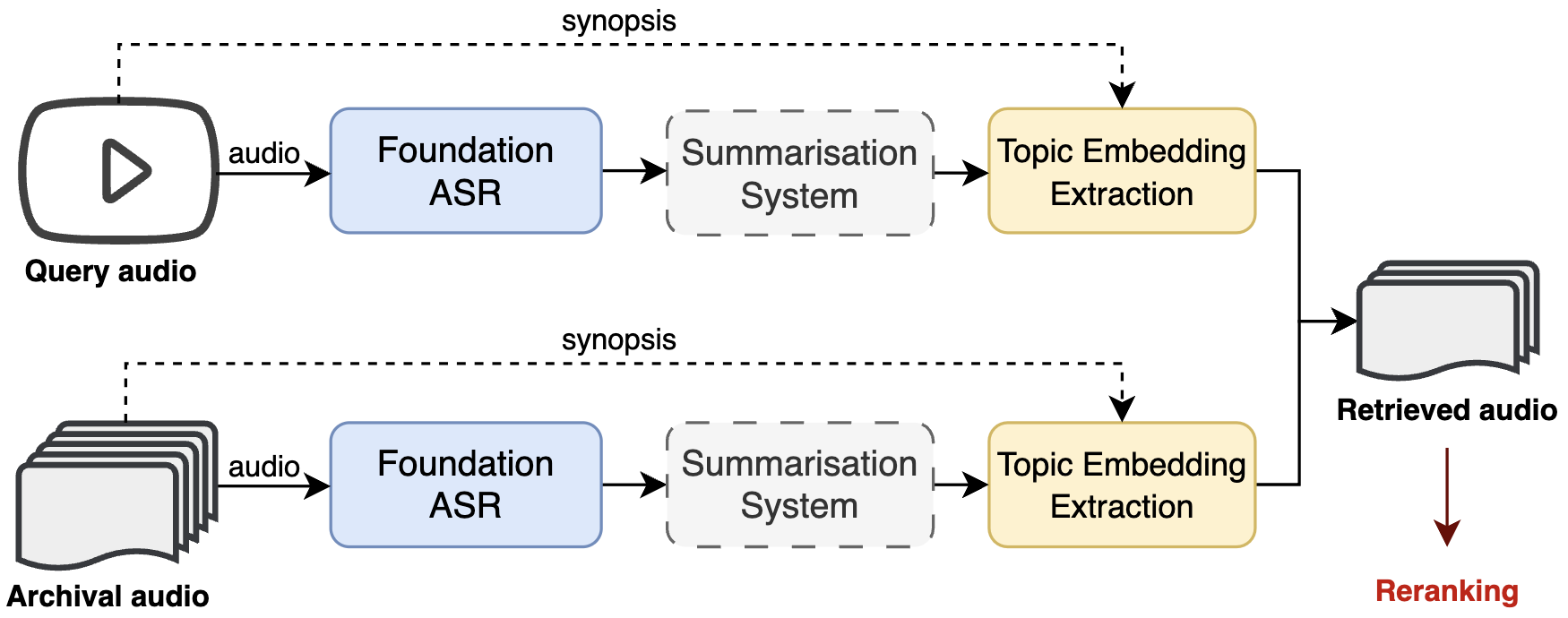}
    \caption{Pipeline of the MVSE audio topic retrieval system.}
    \label{fig:baseline}
    \vspace{-3mm}
\end{figure}

In MVSE, the aim of this work is to accurately extract audio from a large archive that matches the topic of the query. 
All the test queries and each archival audio were annotated with one or more ground-truth topic labels. 
%To enhance the efficiency of retrieval within the context of MVSE, hashing methods are employed to facilitate retrieval across all modalities, as detailed in~\cite{yang2023unsupervised}. Alternatively,
Spoken content retrieval is typically implemented with ASR and a text engine~\cite{lee2015spoken}. 
Alternatively, our baseline approach for audio topic retrieval is based on the embedding space to achieve retrieval across all modalities.
This system follows a multi-step process involving an initial foundation ASR system to transcribe the audio, an (optional) summarisation system, and a pre-trained topic classification model to extract topic embeddings. These embeddings are then used to rank the audio based on cosine similarities between the audio embeddings and the query embedding.
The overall system architecture is illustrated in Figure~\ref{fig:baseline}. With this pipeline, we can develop a retrieval system in the wild without the need for domain-specific training data.
% The developed system allows users to find relevant content based on their topic of interest without the need for in-domain training data. 

In this work, we examine three different types of inputs to extract embeddings for both the query and archival audio:
\begin{enumerate}
    \item \textbf{ASR:} Since the topic embedding extraction model operates on text inputs, an ASR model is used to generate a transcription of the audio. Whisper~\cite{radford2022robust} which shows state-of-the-art performance in the zero-shot setting is used as the underlying ASR model. 
    \item \textbf{AutoSum:} ASR transcriptions can often be lengthy and noisy, especially with audios exceeding 30 minutes. Alternatively, we generate a more concise automatic summary using Llama2-chat~\cite{touvron2023llama} in a zero-shot fashion. A simple prompt is used for this purpose, i.e., \textit{``Write a detailed synopsis of the following transcript:\{transcript\}''}. 
    % Note that this will bring additional costs in the retrieval stage.
    \item \textbf{Synopsis:} As part of the corpus, synopses are crafted by human transcribers with domain expertise, which may include insights from the video content not directly evident in the audio.
    While not the primary focus of our paper, these synopses are included for comparison. It is worth noting that synopses often contain external information that cannot be deduced from the audio content such as speakers' names, locations, and dates, making them richer. In addition, they do not include irrelevant details. The process regarding this is shown in dashed arrows in Figure~\ref{fig:baseline}.
\end{enumerate}

\begin{figure}[!htbp]
    \centering
    \includegraphics[trim=0 5mm 0 5mmmm, width=0.8\linewidth]{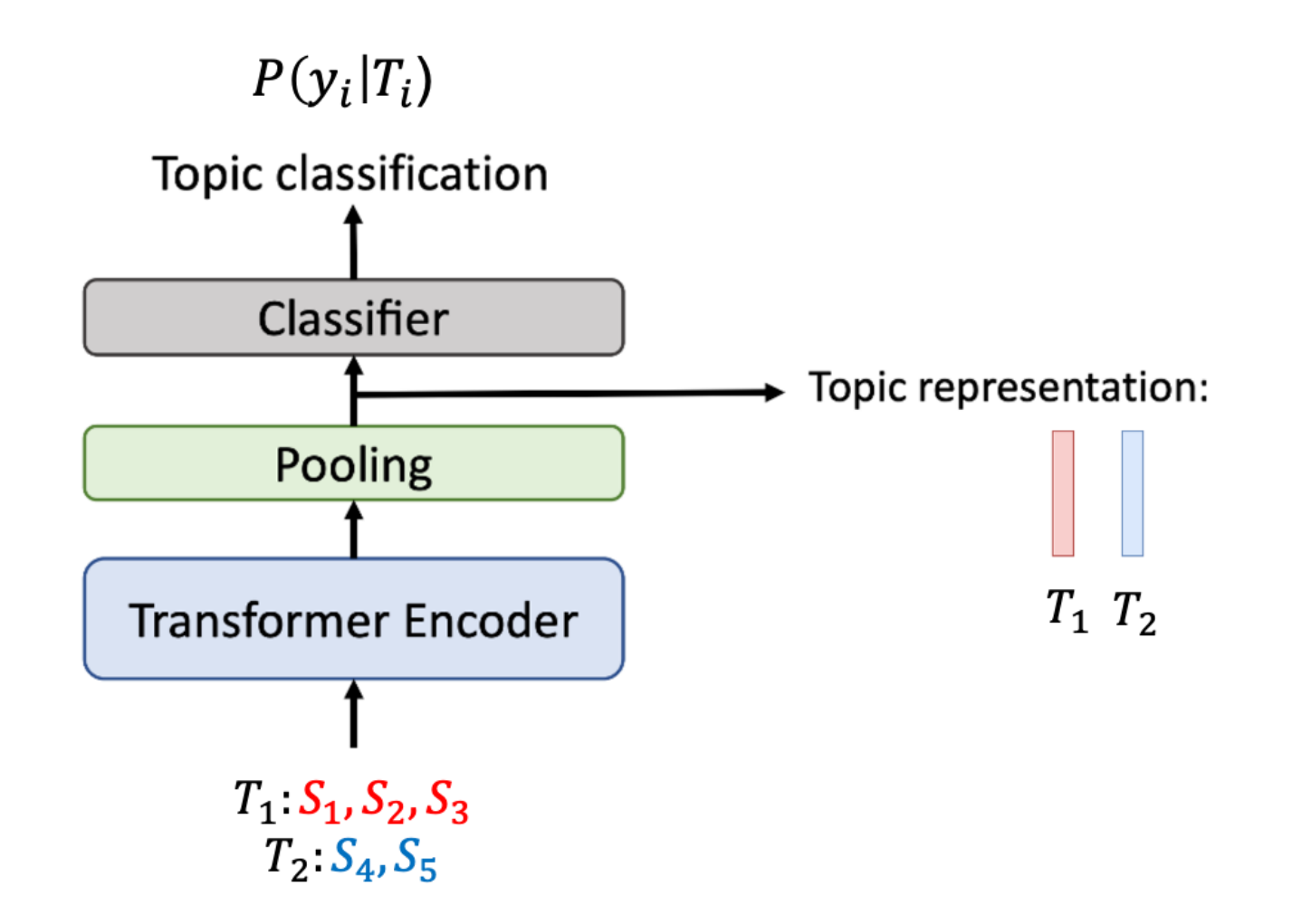}
    \caption{Topic embedding extraction.}
    \label{fig:topic_retrieval}
    \vspace{-3mm}
\end{figure}

Based on these text representations, a novel language model is employed to extract high-quality topic embeddings for retrieval purposes. Given the absence of domain-specific training data, we fine-tune a pre-trained RoBERTa~\cite{liu2019roberta} model on a topic classification task using the publicly available Topic Detection and Tracking (TDT)~\cite{allan1998topic} dataset. The TDT dataset covers a wide range of topics such as Business, Politics, Sport, Entertainment, Science, and Technology. 
Although the topics differ from those in the BBC Rewind corpus, they cover a wide range, allowing the fine-tuned RoBERTa to generalise well to new, unseen topics. Initial experiments compared other pre-trained models, with RoBERTa showing better performance. In the inference phase, vectors from the penultimate layer of the RoBERTa model are extracted to represent topic information, as shown in Figure~\ref{fig:topic_retrieval}.

These extracted embeddings are then used to compute the cosine similarity between the query and each archival audio to perform retrieval on the entire corpus. This pipeline leverages the strengths of pre-trained language models and transfer learning, allowing effective topic extraction and retrieval without the need for domain-specific training data.

\subsection{Information Consistency}
\label{sec:fact}
% %
% \noindent
% Understanding the information consistency between various text inputs used in the retrieval system is crucial.
% %LLMs are prone to generating non-factual statements, which can reduce trust in their output. 
% ``SelfCheckGPT'' offers a simple sampling-based approach to fact-checking LLM responses at the sentence level in a zero-resource fashion \cite{manakul2023selfcheckgpt}. Min et al. proposed a fine-grained two-stage fact-checking method: first, breaking down LLM-generated content into atomic facts, and second, computing the percentage of atomic facts supported by a knowledge source~\cite{min2023factscore}. Their work showed a fact generation and evaluation error rate of less than 2\% compared to human assessment. 
%

\begin{table*}[!htbp]
    \caption{An example of Synopsis, AutoSum and ASR for the same audio file.}
    \label{tab:text}
    \centering
    \footnotesize
     \begin{tabular}{p{10mm}|p{158mm}}
    \toprule
     Source    & Content \\
     \midrule
     \midrule
    Synopsis & \strut{\color{red}\textbf{Nine-year-old} }{\color{blue} \textbf{Emma Good}}'s Book of  {\color{red}\textbf{Prayers `Dear God'}}. Report by {\color{red}\textbf{Maggie Taggart}}. Actuality of `thank you' speech from unidentified member of the {\color{cyan}\textbf{NI Orthopaedic Council}} \\
    \midrule
    AutoSum & \strut{\color{blue}\textbf{Emma}} raised over {\color{orange}\textbf{9,000}} copies of her book for the {\color{cyan}\textbf{Northern Ireland Council for Orthopedic Development}} through sales from {\color{pink}\textbf{all over the world, including Thailand, Germany, Denmark, Canada, and even schools in Mallow and County Cork}}. She is now planning to write {\color{purple}\textbf{a new book of poems}}.\\
    \midrule
    ASR & I'm asking you to write another book? No, I just... There was an effort which you are making on behalf of the {\color{cyan}\textbf{Northern Ireland Council for Orthopedic Development}}. The council will take this check and will use it for a number of purposes. And probably the principal purpose will be to improve the service provided in the Belfast area through the provision of an advice centre at Musgrave Park Hospital. It is a magnificent effort in your part and the council is most grateful which you and your family and all connected with the project have done to produce this magnificent sum. Thank you very much. {\color{blue} \textbf{Emma}}, how did you manage to raise so much money from your book? Well, we had so many books from nearly all over the world and we just thought that it would be nice to raise lots of money so that it would go to the {\color{cyan}\textbf{Northern Ireland Council for Orthopedic Development}}. What sort of countries did you sell your book in? We sold a book away to {\color{pink}\textbf{Thailand and away to Germany and Denmark and Canada and all other countries of the world}}. And some schools bought them as well, didn't they? Yes. Where were they? Everyone was from {\color{pink}\textbf{Mallow and County Court}} and they were very pleased with the book and they used it a lot and almost from the school in Jersey. How many books have you sold, do you know? Over {\color{orange}\textbf{9,000}}. We've almost just reached 10,000 copies. Well, you're going to write a new book, aren't you, when you have time? Yes. What's it going to be about? It is going to be {\color{purple}\textbf{a book of poems}} and I made a poem up and I said to you now, now it's 10 to 3, now come for tea with cakes and bumps for you and me. We'll go by the sea and finish our tea at half past three.\\
    \bottomrule
    \end{tabular}
\end{table*}

\begin{figure}[!htbp]
    \centering
    \includegraphics[trim=0 3mm 0 6mm, width=1.0\linewidth]{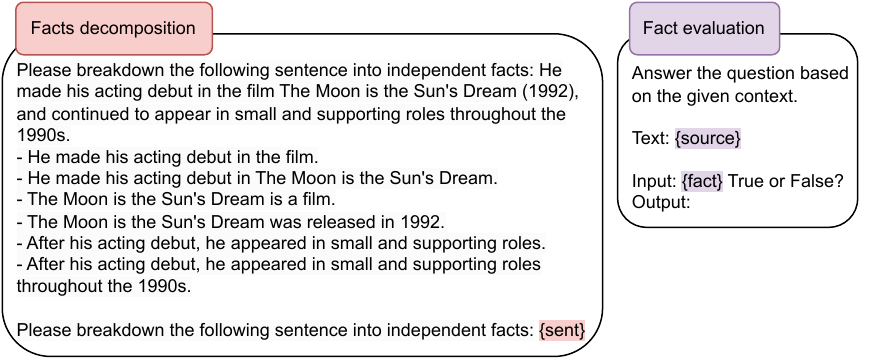}
    % \vspace{-5mm}
    \caption{Prompts for fact checking: 1-shot prompt for facts decomposition and zero-shot for fact evaluation.}
    \label{fig:prompt-fact}
    % \vspace{-3mm}
\end{figure}

\noindent
Both the topic labels and the synopses for the BBC video recordings were annotated by human transcribers. These annotations typically include external information not present in the audio content, such as speakers' names, locations and dates. This additional information, however, can be useful for topic retrieval. 
Table~\ref{tab:text} presents an example with Synopsis, ASR transcriptions and AutoSum. 
Due to space limitations, only one example is shown, but similar patterns are observed in other cases. The synopsis contains the interviewee's age, full name and the interviewer's name, which are not mentioned in the audio content as it only records the conversation between the interviewer and interviewee.
As a result, AutoSum, which is derived from ASR transcriptions, also lacks this information. Analysing the coverage differences between these text sources is essential for understanding the nature of the data and interpreting the experimental results.

A fine-grained two-stage fact-checking method was proposed by Min et al.~\cite{min2023factscore}: first, breaking down LLM-generated content into atomic facts, which are short statements that each contain one piece of information, and second, computing the percentage of atomic facts supported by a knowledge source. Their work showed a fact generation and evaluation error rate of less than 2\% compared to human assessment. 
We apply this fact-checking approach to evaluate the information consistency of \textit{ASR}, \textit{AutoSum}, and \textit{Synopsis} text sources. 
%For example, to verify if the information in AutoSum aligns with ASR, we prompt an LLM to break down each sentence in AutomSum into facts and then check if each fact is supported by ASR transcriptions (source). We use Mistral-7B-Instruct-v0.2~\cite{jiang2023mistral}, an instruct fine-tuned LLM, for this task. The prompts for fact decomposition and fact evaluation are depicted in Figure~\ref{fig:prompt-fact}. A 1-shot prompt suffices for fact decomposition, with the example derived from \cite{min2023factscore}, and the performance remains consistent regardless of the 1-shot example used. The objective is to compare information between different text sources, and a 1-shot prompt effectively demonstrates differences, eliminating the need for larger n-shot evaluations.
For example, to verify if the information in synopses is covered by ASR transcriptions, we prompt an LLM to break down synopses sentences into facts and then check if each fact is supported by ASR transcriptions (source). We use Mistral-7B-Instruct-v0.2~\cite{jiang2023mistral}, an instruct fine-tuned LLM, for this task. The prompts for fact decomposition and fact evaluation are depicted in Figure~\ref{fig:prompt-fact}. A 1-shot prompt, with the example derived from \cite{min2023factscore}, suffices for fact decomposition and the performance remains consistent regardless of the 1-shot example used. %The objective is to compare information between different text sources, and a 1-shot prompt effectively demonstrates differences, eliminating the need for larger n-shot evaluations.
This approach compares information between different text sources effectively without needing larger n-shot evaluations.
Fact evaluation is conducted in \textit{free generation} mode, allowing the model to generate options beyond ``true'' or ``false'', such as ``i cannot decide based on the given context''. A detailed explanation for the \textit{free generation} mode is included in Section~\ref{sec:pairwise}.

% We are using LLMs to generate summaries for ASR transcriptions, it's important to understand if the generated summary has non-factual statements. 

% Describe the method for fact checking, and other similar approaches, examine this in the experiments

\section{Zero-shot Reranking Methods}
\label{sec:method}
\noindent
Due to the lack of task-specific data and the challenge of transcribing noisy audio accurately, MVSE audio topic retrieval performance can be restricted. In this study, we explore whether employing zero-shot reranking methods with powerful LLMs as a second stage can boost retrieval performance. Specifically, we perform reranking on the Top $N$ retrieved results. In line with the methods listed in Section \ref{sec:mvse}, three types of text input can be used as input for the reranking methods. We examine the use of \textit{ASR}, \textit{AutoSum}, or \textit{Synopsis} as queries or documents in the experimental section. 
%For summarisation and reranking, we deliberately kept the prompts simple and avoided overengineering to maximise the practicality.

\subsection{Listwise Reranking}
\begin{figure}[!h]
    \centering
    \includegraphics[trim=0 3mm 0 3mm, width=0.86\linewidth]{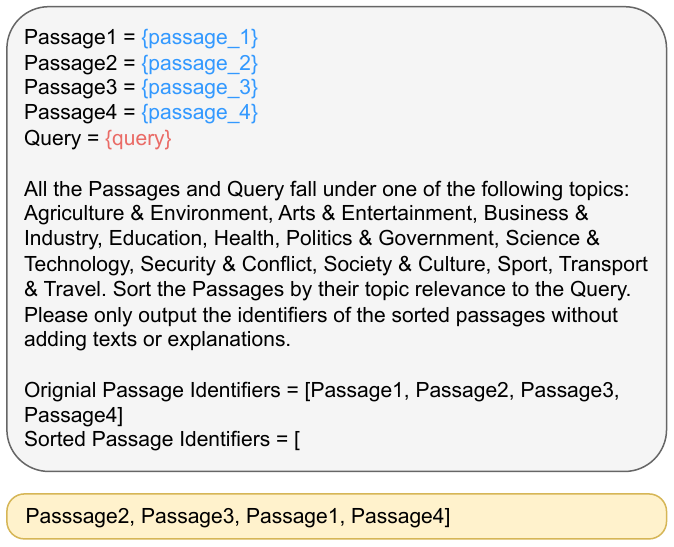}
    \caption{Prompt design for listwise topic reranking.}
    \label{fig:listwise}
    % \vspace{-1mm}
\end{figure}
\noindent
In \cite{ma2023zero, sun2023chatgpt}, a zero-shot listwise reranking method with LLMs (referred to as LRL) is adopted. The method accepts a query and a list of candidate documents as input and aims to generate a reordered list of document identifiers based on their relevance to the query. With this approach, the model can attend to information regarding all candidates and is expected to generate the reranked order in one go. The prompt design for listwise audio topic reranking is shown in Figure \ref{fig:listwise}, where the LLM is instructed to compare and rerank all $N$ candidates based on the topic relevance to the given query text. In experiments, documents are listed in the same order as the initial retrieval outputs. Our preliminary experiments show that similar performance can be obtained on the test set regardless of employing a random or reverse order.%when employing a random or reverse order.

\subsection{Pairwise Reranking}
\label{sec:pairwise}
\begin{figure}[!htbp]
    \centering
    \includegraphics[trim=0 5mm 0 0mm, width=0.9\linewidth]{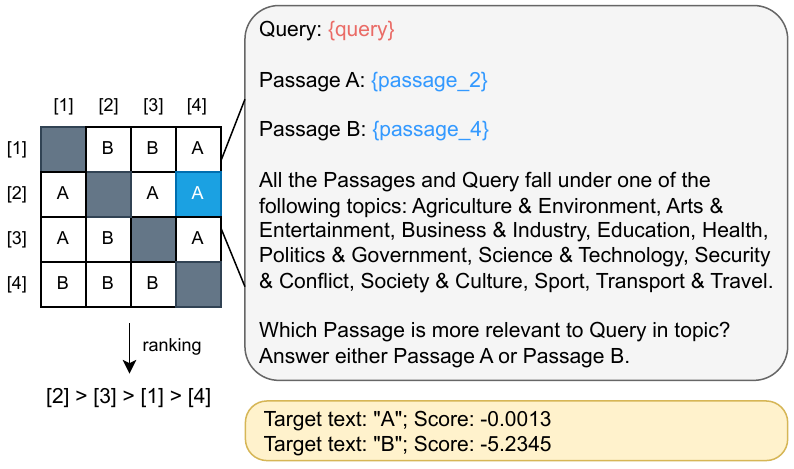}
    \caption{Prompt design for pairwise topic reranking.}
    \label{fig:pairwise}
    \vspace{-5mm}
\end{figure}

\noindent
While LRL demonstrates promising performance in certain tasks \cite{ma2023zero}, a significant concern arises when dealing with a large value of $N$, where the total token count of the concatenated documents will exceed the limit set by LLMs. Additionally, it can be difficult for the model to compare all the given candidates to generate the absolute reranked order. 
To address this, alternative approaches such as reranking the entire list with a sliding window strategy have been adopted \cite{sun2023chatgpt}. %\cite{qin2023large, liusie2023zero} considered the pairwise ranking using LLMs paradigm (PRL). Pairs of responses are compared in each round to determine which one is considered better. 
Pairwise ranking using LLMs (PRL) was studied in \cite{liusie2024efficient, liusie2024teacher}, where pairs of responses are compared in each round to determine which one is considered better. For a candidate list of $N$ documents, $N*(N-1)$ comparisons can be computed, with the win ratio used to rank the $N$ documents.

In Figure~\ref{fig:pairwise}, we illustrate the prompt-based classification method used in our PRL experiments. In general, there are two possible decoding strategies: \textit{free generation} and \textit{prompt-based} classification. 
In the former, the model freely generates decoded text sequences that are then mapped into either Passage A or Passage B. Since the LLM has the freedom to generate diverse texts, complex rules may be required to map the texts to the desired outcomes. 
% Alternatively, prompt-base classifiers \cite{liusie2023mitigating} compare the likelihood of the model generating text \textit{``A''} or \textit{``B''} as shown in Figure \ref{fig:pairwise}. This approach avoids an expensive beam search while also simplifying the decoding approach. Our initial experiments show that the two methods perform similarly, therefore, the low-cost prompt-based classification is used in this paper.
Alternatively, prompt-based classifiers \cite{liusie2023mitigating} simplify decoding by comparing model likelihoods to generate \textit{``A''} or \textit{``B''}  as shown in Figure \ref{fig:pairwise}, avoiding complex rules and expensive searches. Initial tests indicate both methods perform similarly, so we use the low-cost prompt-based classification.

%Observing the problem of positional bias in pairwise comparisons, ~\cite{liusie2023zero} proposed debiasing methods and obtained state-of-the-art performance for zero-shot natural language generation (NLG) assessment tasks.
%the scoring mode, we only calculate and compare the score that the model generates certain outputs. Our initial experiments show that the two methods perform similarly on the test set and the scoring mode is selected in this paper since it is more efficient. To be specific, we compare the 

\section{Experimental Results}
\label{sec:exp}
\subsection{Data Setup}  
\begin{figure}[!htbp]
    \centering
    \includegraphics[trim=0 5mm 0 5mm,width=0.9\linewidth]{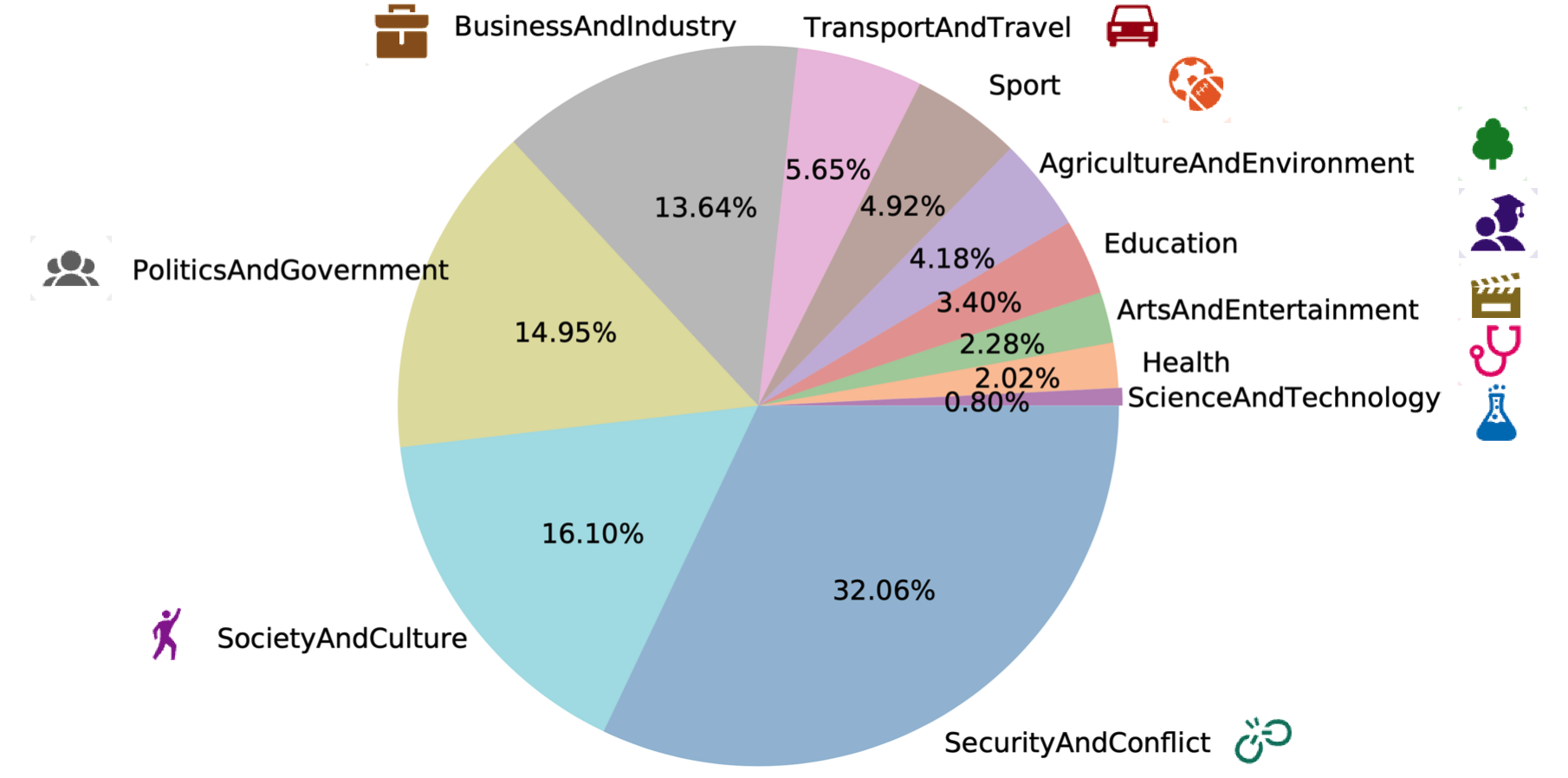}
    \caption{Topic distribution in the BBC Rewind dataset.}
    \label{fig:topic_distribution}
    \vspace{-2mm}
\end{figure}

\noindent
The audio recordings associated with the public BBC Northern Ireland Rewind audio corpus\footnote{https://bbcrewind.co.uk/} are used in the experiments. After filtering audio with no speech content, the archive consists of 7855 audio documents with an average duration of 117 seconds. These audio recordings vary greatly in length, ranging from 3 seconds to over 30 minutes.
Each audio is associated with one or more topic labels according to the meta-data, such as AgricultureAndEnvironment, Education, and Health. 
There are 11 different topics, and their distribution in the recordings is illustrated in Figure~\ref{fig:topic_distribution}.
We conduct extensive experiments using this diverse audio dataset, randomly selecting 107 audio clips from this dataset as the query list. Each clip is attributed to a single topic, ensuring a comprehensive evaluation of our reranking approach. 

For the foundation ASR system, we adopt Whisper small.en model, which shows comparable performance to larger ones with less computational cost.
% In our experiments, we use nDCG@\{3,5,10\} and precision@\{1,3,5\} as evaluation metrics \cite{jarvelin2002cumulated}, all of them are the higher the better. 
% In our experiments, we use normalized discounted cumulative gain(nDCG)\cite{jarvelin2002cumulated}, a metric measures the ranking quality of the output list compared to the ideal ranking, and Precision@k, which measures the proportion of relevant items among the top k items in the output list. Specifically, we used nDCG@\{3,5,10\} and Precision@\{1,3,5\}.
%
For evaluation, normalized discounted cumulative gain (nDCG) \cite{jarvelin2002cumulated,ndcg2013} and Precision@k metrics are employed. The nDCG measures the ranking quality compared to the ideal ranking, while Precision@k measures the proportion of relevant items among the top-k items in the output list. Specifically, we use nDCG@\{3,5,10\} and Precision@\{1,3,5\}. Note that nDCG@1 is equivalent to Precision@1, so it was not included separately.
If not stated otherwise, we perform reranking on the Top-10 audios retrieved from the first stage. The model performance with different sizes of the candidate list is studied in the ablation.

% Topic list: agricultureAndEnvironment
% artsAndEntertainment
% businessAndIndustry
% education
% health
% politicsAndGovernment
% scienceAndTechnology
% securityAndConflict
% societyAndCulture
% sport
% transportAndTravel

% \begin{equation}
%     \text{NDCG} = \frac{\text{DCG}@K}{\text{IDCG}@K}
% \end{equation}
% \begin{equation}
%     \text{DCG@K} = \sum_{k=1}^K \ \frac{\text{rel}_k}{\log(1+k)}
% \end{equation}

\subsection{First-stage Retrieval}
In the first-stage, the audio topic retrieval system returns 1000 candidates and Table~\ref{tab:baseline} shows the results using three text sources to extract topic embeddings. There is a big performance gap when using ASR transcriptions compared to manual synopses as input. The system built from \textit{AutoSum} of ASR transcriptions is also examined. With this modification, we hope to generate automatic summaries that reduce noise in ASR transcriptions and mimic the behaviour of human-generated synopses. As shown in Table~\ref{tab:baseline}, the retrieval system built with AutoSum outperforms the one using the original ASR transcription. Nevertheless, this will add computational overhead compared to using the ASR outputs directly. Despite this improvement, the \textit{AutoSum} system still performs worse than the \textit{Synopsis} system. This is likely due to the synopses having richer content as the humans can add information present in the video but not in the audio which the ASR transcriptions and AutoSum rely on. A detailed analysis of this is provided in Section~\ref{sec:exp_fact}, where we found that synopses contain more than 50\% atomic facts not included in ASR. The example in Table~\ref{tab:text} also illustrates it's typical for synopses to include information not present in the audio content.

\begin{table}[!t]
    \caption{Baseline retrieval performance when different text sources are used for embedding extraction.}
    \label{tab:baseline}
    \centering
    \small
    \begin{tabular}{l|ccc|ccc}
    \toprule
        \multirow{2}{*}{Retrieval} & \multicolumn{3}{c|}{nDCG} & \multicolumn{3}{c}{Precision} \\
        & @3 & @5 & @10 & @1    & @3    & @5    \\
    \midrule
        ASR  &  0.47 & 0.46 & 0.45 &  0.48 & 0.47 & 0.45  \\
        AutoSum  & 0.52 & 0.50 & 0.48 & 0.57 & 0.51 & 0.48  \\
        Synopsis & 0.61 & 0.59 & 0.55 & 0.62 & 0.60 & 0.57  \\
        \bottomrule
    \end{tabular}
    \vspace{-2mm}
\end{table}

\begin{table}[!t]
    \caption{Oracle performance of Top 10 retrieval results.}
    \label{tab:oracle}
    \centering
    \small
    \begin{tabular}{l|cccc}
    \toprule
        Retrieval & P@1 & P@3 & P@5 \\
    \midrule
        % ASR & 0.86 & 0.74 & 0.62 & 0.42 \\
        ASR & 0.85 & 0.74 & 0.62 \\ 
        AutoSum & 0.92 & 0.78 & 0.67 \\
        Synopsis &  1.00 & 0.89 & 0.78 \\
        \bottomrule
    \end{tabular}
    \vspace{-2mm}
\end{table}

The oracle precision for the Top 10 retrieved candidates from different retrieval systems is listed in Table \ref{tab:oracle}. As the oracle nDCG is always 1 by definition, it is not included. The results indicate that the upper bound for the reranking task is quite high, suggesting potential improvements by incorporating reranking as a second stage using powerful LLMs.

\subsection{Experiments on Zero-shot Reranking}
% In Table~\ref{tab:compare_rank}, we take the retrieval system based on ASR transcriptions and compare different zero-shot reranking methods.
Table~\ref{tab:compare_rank} compares zero-shot reranking methods for the retrieval systems based on ASR transcriptions.
BERTScore~\cite{Zhang2020BERTScore} and the ROUGE score~\cite{lin-2004-rouge} are commonly used metrics to measure the similarity between two texts.
These metrics can also be used to measure the similarity between a query text and the archive texts for ranking purposes. 
Our experimental results indicate that using BERTScore and ROUGE scores for reranking did not yield any performance gain. This could be attributed to the fact that BERTScore is primarily designed to evaluate the quality of a machine-generated test against a reference, rather than being inherently tailored for ranking tasks.

\begin{table}[!t]
\caption{Reranking performance with different methods and LLMs using ASR transcriptions as input.}
    % \vspace{-2mm}
    \label{tab:compare_rank}
    \centering
    \small
    % \begin{tabular}{@{ }l|l@{ }|ccc|ccc@{ }}
    \begin{tabular}{@{ }l@{ }|@{ }l@{ }|C{4.6mm}C{4.6mm}C{5.4mm}|C{4.6mm}C{4.6mm}C{4.6mm}}
    \toprule
    \multicolumn{2}{c|}{\multirow{2}*{Rerank}} & \multicolumn{3}{c|}{nDCG} & \multicolumn{3}{c}{Precision} \\
    \multicolumn{2}{c|}{ } & @3 & @5 & @10 & @1    & @3    & @5    \\
    \midrule
     % \multicolumn{2}{l|}{Baseline} & 0.61 & 0.59 & 0.55 & 0.62 & 0.60 & 0.57  \\
     % \midrule
     % gpt-3.5-turbo-16k
        \multicolumn{2}{l|}{Baseline} & 0.47 & 0.46 & 0.45 &  0.48 & 0.47 & 0.45   \\
        \midrule
        \multicolumn{2}{l|}{BERTScore}& 0.42 & 0.43 & 0.28 & 0.45 & 0.41 & 0.43 \\
        \multicolumn{2}{l|}{ROUGE} &0.42 & 0.42 & 0.27 & 0.40 & 0.42 & 0.43\\
     \midrule
        \multirow{3}{*}{\rotatebox{90}{LRL}} & gpt-3.5 & 0.48 & 0.47 & 0.46 & 0.49 & 0.47 & 0.47 \\
        & gpt-4 & \bf 0.52 & \bf 0.50 & \bf 0.47 & \bf 0.59 & \bf 0.50 & 0.48 \\
        & Flan-T5-3B  & 0.45 & 0.45 & 0.45 & 0.47 & 0.45 & 0.44 \\
    \midrule
        \multirow{4}{*}{\rotatebox{90}{PRL}} & Llama2-chat-7B & 0.45 & 0.45 & 0.30 & 0.45 & 0.45 & 0.46\\
        % & Llama2-chat-13B & \\
        & Flan-T5-220M & 0.46 & 0.44 & 0.29 & 0.45 & 0.46 & 0.44 \\
        & Flan-T5-770M &  0.50 & 0.48 & 0.31 & 0.52 & 0.50 & 0.46\\
        & Flan-T5-3B  & \bf 0.52 & \bf 0.50 & \bf 0.47 & \bf 0.59 & \bf 0.50 & \bf 0.49\\
        % & Flan-T5-11B & \\
        \bottomrule
    \end{tabular}
    \vspace{-2mm}
\end{table}

The results of zero-shot reranking using LLMs are listed in Table \ref{tab:compare_rank}. For the LRL method, adopting the GPT-3.5 (\texttt{gpt-3.5-turbo-16k}) model yields similar performance to the baseline system, while GPT-4 (\texttt{gpt-4-0613})~\cite{openai2023gpt4} shows improved performance. Open-source LLMs such as Flan-T5~\cite{chung2022scaling,raffel2020exploring} do not improve the performance, likely due to their limitations on maximum token input.
For the PRL method, we employed advanced open-source LLMs to obtain necessary likelihood scores, as GPT models do not provide such scores\footnote{GPT models in recent release can provide likelihood scores, but this feature was not available when these experiments were conducted.}.
Results show Flan-T5 is more effective than Llama2-chat~\cite{touvron2022llama,touvron2023llama} in assessing the topic relevance of candidates to the query. Since PRL involves shorter context inputs compared to LRL, Flan-T5 performs well for PRL even though it does not show performance gains for LRL.
Both GPT and Flan-T5 demonstrate improved performance with larger model sizes across all metrics.
The best-performing PRL method shows similar performance to the LRL system while a foundation model with far fewer parameters is required. 
Reranking significantly enhances overall system performance, with a relative improvement of 10.6\% on nDCG@3 and 22.9\% on Precision@1. 
In the following experiments, we utilise PRL as the main reranking method and Flan-T5-3B as the underlying model.

\begin{table}[!t]
    \caption{Reranking performance with PRL with different $N$.}
    % \vspace{-3mm}
    \label{tab:size}
    \centering
    \small
    \begin{tabular}{l|ccc|ccc}
    \toprule
        \multirow{2}*{$N$} & \multicolumn{3}{c|}{nDCG} & \multicolumn{3}{c}{Precision} \\
        & @3 & @5 & @10 & @1    & @3    & @5    \\
    \midrule
    %  Baseline & - & 0.47 & 0.46 & 0.45 &  0.48 & 0.47 & 0.45    \\
    % \midrule
        5 & 0.49 & 0.47 & 0.31 & 0.53 & 0.48 & 0.45\\
        10 &  0.52 & 0.50 & 0.47 & 0.59 & 0.50 & 0.49\\
        20 & 0.52 & 0.50 & 0.32 & 0.53 & 0.51 & 0.49\\
        \bottomrule
    \end{tabular}
    \vspace{-2mm}
\end{table}

The ablation on different sizes of the candidate list (N) is shown in Table \ref{tab:size}. From the results, we can see that there is improvement when we increase $N$ from 5 to 10 while the performance degrades when the list is further expanded to 20. This indicates that increasing $N$ improves performance up to a point, but a larger $N$ can overwhelm the model and reduce effectiveness.

% Next, we explore the usage of different text sources in the reranking stage. 
%For the retrieval system built from ASR transcriptions, when we use the summarised ASR transcriptions as input to LLMs, it brings a small performance gain compared to using the ASR transcription as input. However, for the retrieval system built from ASR summaries, changing the input from ASR transcriptions to the auto-summaries shows slightly worse performance. 
% In this case, we do not exploit the richer context in the ASR transcriptions during both retrieval and reranking stages. 
% The loss of information can be harmful for the model to decide on the topic of the audio. 
Next, we explore the usage of AutoSum for improving reranking performance. Reranking with ASR transcriptions demonstrated a consistent performance improvement over baseline retrieval. Given this, reranking using AutoSum further improves the performance, with nDCG@3 increasing from 0.52 to 0.54 and Precision@3 from 0.50 to 0.52, as shown in Table~\ref{tab:compare_results}.
Changing the text input from ASR transcriptions to AutoSum at the retrieval stage shows a similar performance gain, indicating that a cleaner text input can yield a better ranking performance.
% Additionally, zero-shot reranking based on synopses metadata proves effective, showcasing the continued efficacy of reranking when additional audio metadata is utilized. Leveraging named entities annotated in synopses with Large Language Models (LLMs) notably enhances system performance, underscoring the potential of LLM-reranking. 
In addition, we show the performance of zero-shot reranking for the system based on synopses. The results indicate that reranking continues to be effective when meta-data of the audio is available, underscoring the potential of LLM reranking. 
However, we cannot use synopses for practical applications due to the limited availability of synopses for system development.

\begin{table}[!t]
    \caption{Retrieval and reranking using different text sources.}
    \label{tab:compare_results}
    \centering
    \small
    \begin{tabular}{@{ }l@{ }|@{ }l@{ }|C{4.5mm}C{4.5mm}C{4.5mm}|C{4.5mm}C{4.5mm}C{4.5mm}}
    \toprule
        \multirow{2}{*}{Retrieval} & \multirow{2}{*}{Rerank}  & \multicolumn{3}{c|}{nDCG} & \multicolumn{3}{c}{Precision} \\
        & & @3 & @5 & @10 & @1    & @3    & @5    \\
    \midrule
    % \midrule
        \multirow{1}*{ASR} & - & 0.47 & 0.46 & 0.45 &  0.48 & 0.47 & 0.45  \\
        % \cmidrule{2-8}
         ASR & ASR & 0.52 & 0.50 & 0.47 & 0.59 & 0.50 & 0.49\\
         \midrule
         ASR & AutoSum & 0.54 & 0.51 & 0.47 & 0.58 & 0.52 & 0.49\\
         % & Synopsis & 0.61 & 0.57 & 0.49 & 0.69 & 0.58 & 0.53\\
        \multirow{1}*{AutoSum} & ASR & 0.54 & 0.52 & 0.49 & 0.59 & 0.52 & 0.51 \\
        \midrule
        % \cmidrule{2-8}
        \multirow{1}*{Synopsis} & Synopsis & 0.73 & 0.68 & 0.59 & 0.76 & 0.71 & 0.65 \\
        \bottomrule
    \end{tabular}
    \vspace{-2mm}
\end{table}

\subsection{Analysis on Information Consistency}
\label{sec:exp_fact}
We hypothesise that the performance gap between \textit{ASR} transcriptions and \textit{Synopsis} arises because synopses often contain external information not present in the original audio, which ASR transcriptions might not capture. To test this, we adopt the fact-checking approach detailed in Section~\ref{sec:fact} to assess information consistency.
This approach is also used to evaluate the quality of \textit{AutoSum}, checking for misleading information introduced during the summary generation stage. Experiments are conducted on a subset of 500 files randomly selected from the archive.
As shown in Table~\ref{tab:factscore}, \textit{Synopsis} has only 47\% of facts supported by ASR, with 18.6\% facts not supported and 34\% undecidable. This indicates that synopses provide richer information compared to ASR transcriptions, supporting our hypothesis.
\textit{AutoSum} only has 3\% facts not supported by \textit{ASR}, with an 87.5\% support rate, suggesting that \textit{AutoSum} is a clean summary of ASR transcriptions. 
When breaking down ASR sentences into atomic facts, we found that ASR transcriptions produce 6-9 times more atomic facts than those in \textit{Synopsis} or \textit{AutoSum}, highlighting the noisy nature of ASR outputs. Consequently, it's difficult for \textit{Synopsis} or \textit{AutoSum} to cover all atomic facts present \textit{ASR}.
Despite \textit{AutoSum} being cleaner and more concise compared to \textit{ASR}, Table~\ref{tab:factscore} also reveals a mismatch between \textit{Synopsis} and \textit{AutoSum}, suggesting that it is challenging for \textit{AutoSum} to achieve similar performance as \textit{Synopsis}.

\begin{table}[!t]
    \caption{Fact checking for a subset of 500 files randomly selected from the archive. ``Other": In most cases the model can not decide whether True or False based on the input.}
    \label{tab:factscore}
    % \small
    \centering
    \begin{tabular}{@{ }cc@{ }|C{8mm}C{8mm}C{8mm}C{8mm}}
    \toprule
    Facts & Source & \#facts & \%True & \%False & \%Other \\
    \midrule
    Synopsis  & ASR & 3226 & 46.99 & 18.57 & 34.44 \\
    AutoSum & ASR & 2257 & 87.51 & 2.97 & 9.53 \\
    ASR &  Synopsis  & 19346 & 26.55 & 19.97 & 53.48\\
    ASR & AutoSum & 19532 & 31.61 & 27.02 & 41.37\\
    Synopsis & AutoSum &  3226 & 26.04 & 28.46 & 45.51\\
    AutoSum & Synopsis & 2257 & 43.73 & 17.50 & 38.77\\
    \bottomrule
    \end{tabular}
    \vspace{-0mm}
\end{table}

\section{Conclusions}
% In this paper a reranking approach is described that aims to mitigate any performance degradation from supporting very rapid information retrieval from video archives using video queries, in particular for topic retrieval. 
This paper presents a robust two-stage audio topic retrieval system, where the first stage operates on the embedding space to facilitate rapid information retrieval from video archives, and the second stage employs a reranking approach to address potential performance degradation.
% when facilitating rapid information retrieval from video archives using video queries, in particular for topic retrieval. 
A critical challenge for the reranking is to develop approaches that can operate on a wide range of video archives motivating the use of zero-shot approaches based on LLMs. Two forms of representation of the audio, which are then converted into embeddings, are examined: raw ASR output; and the use of an LLM-based ASR summarisation stage to remove redundant and irrelevant information. %Performance for topic retrieval is evaluated on a publicly available dataset, the BBC Rewind corpus, where the use of summarisation and LLM-ranking show gains over the baseline retrieval approach, and other zero-shot reranking.
Evaluation on the BBC Rewind corpus demonstrates that incorporating summarisation and LLM-based reranking yields significant improvements over baseline retrieval methods and other zero-shot approaches.
In the future, we will apply the proposed methods to audio retrieval for other attributes such as content or emotion.

% \section{ACKNOWLEDGMENTS}
% \label{sec:ack}

% Do not include acknowledgments in the initial version of the paper submitted for blind review.
% If a paper is accepted, the final camera-ready version can (and probably should) include acknowledgments. 

\newpage
\bibliographystyle{IEEEbib}
\bibliography{strings,refs}

\end{document}